%% file: main.tex
\documentclass{article}
\usepackage[dblblindworkshop,preprint]{neurips_2026}
\workshoptitle{Agentic AI for Biological Discovery}
\usepackage{amsmath,amssymb,booktabs,graphicx,float}
\title{When Does Feedback Help?\\Auditing Learned Voltage-Clamp Experimental Design}
\author{
  Rayan Nyaupane  \\
  The University of Texas at Austin \\
  \texttt{rayannyaupane@utexas.edu} \\
  \And
 Abhiram Bhupatiraju \\
   The University of Texas at Austin \\
  \texttt{ab82836@my.texas.edu} \\
}
\begin{document}
\maketitle
\begin{abstract}
A learned scientific controller can outperform a misspecified planner without
benefiting from feedback. We audit two synthetic voltage-clamp tasks with a
training-selected fixed protocol, current masking, a Bayesian reference, and an
excluded target. Five recurrent policies repeat one fixed protocol in all
1,200 off-grid rollouts; masking current changes no decision. In a crossed
four-hypothesis task, a fixed diagnostic pulse is followed by one learned
choice. Policies then use the first current, but tie a correctly specified
Bayesian controller, and a fitted threshold reproduces the learned actions
without being that controller. The excluded-target prediction gain is
$4.6991\times10^{-6}$ nats per sample. Feedback use and protocol quality are
separate tests. These cases establish neither biological discovery nor general
RL superiority.
\end{abstract}

\section{Question and contribution}
Comparing a learned controller with an online planner mixes the observation
model, the objective, intervention cost, the horizon, and whether outcomes
change later actions. Protocol quality and feedback use are different tests: a
controller that repeats a good fixed experiment has not shown that feedback
improves discovery. We keep three questions separate. Does feedback change
actions? Does it improve an independently scored outcome? Does learning
outperform a simpler adaptive choice?

We apply four controls on two small voltage-clamp libraries: a training-selected
fixed protocol, current masking, a correctly specified Bayesian reference, and
an excluded prediction target. The first setting trains on distorted recordings
and appears to beat ideal-clamp planners. The second rewards a different second
pulse in different latent regimes. This is an empirical audit, not a new
algorithm. Related amortized and reinforcement-learning design includes Deep
Adaptive Design~\citep{foster2021,ivanova2021,hedman2025}, sequential
RL~\citep{blau2022,shen2025}, biological experimental design~\citep{treloar2022},
and waveform optimization for channel discrimination~\citep{kargol2013}.

\section{Original task and prespecified audit}
\paragraph{Cell, actions, and training.}
A synthetic three-state population, not a fitted real cell, follows
$\dot p=pQ(V_m)$ with rates $k_j(V)=A_j\exp(s_jB_jV)$ and current
$I_{\mathrm{ion}}=g p_O(V_m-E_{\mathrm{rev}})$. Occupancy persists between pulses.
The distorted clamp uses $R_s=20$~M$\Omega$ and $C_m=40$~pF, without compensation,
leak, pipette capacitance, or offset, in a circuit motivated by \citet{lei2020}.
A five-point log grid spans a factor of ten around the activation prefactor
$A_1$; other kinetics are known. Action S is 2~ms at $+40$~mV and action L is
12~ms at 0~mV, each with a 2~ms post-step at $-80$~mV. Two actions give durations
of 8, 18, or 28~ms. That imbalance belongs to this task: comparisons match
intervention count, not recording time. Equal recording time applies only inside
the later crossed task. A GRU with eight hidden units sees current, command,
mask, and elapsed time. Training is the original REINFORCE rule: 40 episodes
after 12 baseline-only warmup episodes, Adam step size 0.2, entropy coefficient
0.05, and a moving baseline of rate 0.1. All seeds 0--4 are kept. The reward is
predictive log-likelihood gain on a 6~ms $+20$~mV pulse that also entered
training, so rescoring it is an optimization check. Belief updates use the true
training apparatus and 50~pA Gaussian noise; training traces are noise-free.

\paragraph{Controllers and held-out grid.}
We compare the five frozen policies with the training-selected fixed sequence,
random actions, ideal-clamp information planners, and oracle-apparatus planners.
Greedy maximizes expected parameter-entropy reduction. Lookahead enumerates
remaining sequences, executes the first action, and replans; its horizon is the
number of decisions left, two and then one, not a full contingent tree. Oracle
planners know the test apparatus. A separate oracle predictive planner optimizes
the original training target and never sees the new targets. The audit uses 256
posterior-predictive draws. Before seeing outcomes we fixed
$A_1/A_1^{\mathrm{train}}\in\{0.7,1.4,2.2\}$,
$(R_s,C_m)\in\{(0,0),(10,20),(20,40),(30,60)\}$, noise standard deviations 0 and
50~pA, and recording seeds 100--109: twelve cells and 240 combinations per
controller or training seed. Off-grid kinetics are absent from the inference
grid. Test pulses at $-20$, $+10$, and $+40$~mV never enter training and start
from rest. A shared scorer uses the true apparatus, isolating design quality
under oracle inference. Masking removes only current, not timing or voltage.

\section{Results}
\begin{figure}[t]
\centering
\includegraphics[width=\linewidth,height=1.55in,keepaspectratio]{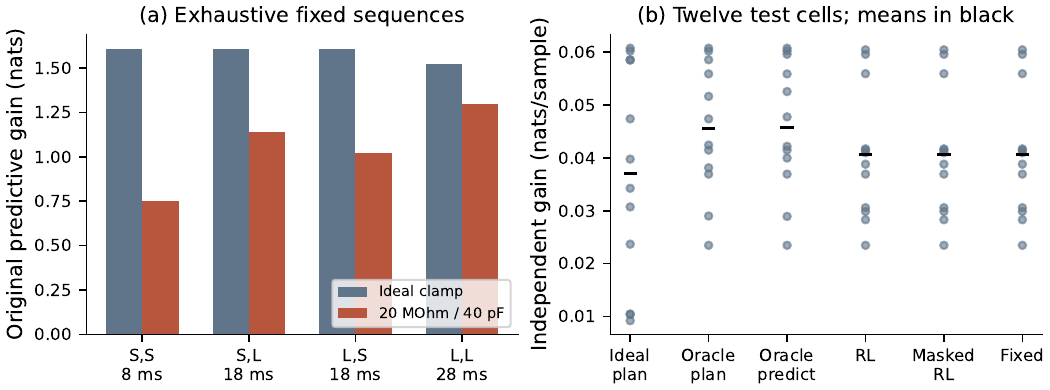}
\caption{(a) Fixed sequences on the original reward, including post-steps.
(b) Independent gains within twelve cells. Points are designed conditions.
RL, masked RL, and the training-selected fixed protocol coincide.}
\label{fig:audit}
\end{figure}

The learned policies look favorable only because they repeat a strong fixed
protocol. On the distorted training cell, SS, SL, LS, and LL score 0.7543,
1.1379, 1.0206, and 1.2986 nats. All five policies choose LL, the
training-selected sequence. On the ideal cell LL scores 1.5240, below SS at
1.6094. LL records for 28~ms, SS for 8~ms, and either mixed sequence for 18~ms.
We make no equal-time claim for this 8/18/28~ms imbalance.
Across all 1,200 rollouts (five seeds times 240 combinations), at both 0 and
50~pA, every policy chooses LL. Masking preserves all 1,200 sequences. That does
not prove the network ignores every possible input. These policies had limited
training on a fixed cell: 40 REINFORCE episodes after 12 warmup episodes, one
apparatus, and a five-point grid on one prefactor. Failure to use feedback here
does not establish that every well-trained recurrent policy would do the same.
Oracle planners in Table~\ref{tab:independent} have higher mean predictive gain.
RMSE and likelihood do not rank methods identically.
The historical runner valued an unexecuted final measurement. The audit caps
the horizon by remaining decisions; the supplement compares capped and uncapped
plans. This correction changes a planner control, not the frozen policies.

\begin{center}
\noindent\begin{minipage}{\linewidth}
\centering
\refstepcounter{table}\label{tab:independent}
{\small Table~\thetable: Independent means on the prespecified grid. Gain is
nats per sample, RMSE is pA, and duration is ms. Oracle methods know the
apparatus.}\par\vspace{3pt}
\footnotesize
\setlength{\tabcolsep}{5pt}
\begin{tabular}{lrrr}
\toprule
Controller & Gain $\uparrow$ & RMSE $\downarrow$ & Duration \\
\midrule
\input{results_table.tex}
\end{tabular}
\end{minipage}
\end{center}

\section{A positive control}
The original audit lacks a regime where feedback should help. We construct four
synthetic hypotheses, not a fitted recording, on the same three-state channel
under an ideal clamp. Two regimes differ in inactivation. Activation varies in
the first and recovery in the second, so the useful second pulse can depend on
the first current. At nominal noise the diagnostic pulse separates the regimes.
Before freezing the task we retained three training-only pilots. Each either
saturated identification or gave no benefit to a regime-informed second choice.
Every method applies the same 30~ms diagnostic pulse and then chooses one of
four equal-duration 30~ms pulses, so all methods record for 60~ms. The decision
structure is a fixed diagnostic pulse followed by one learned choice. It tests
observation-conditioned selection. It does not require long-horizon credit
assignment or recurrent memory. Four 32-unit categorical policies train for
1,600 episodes. The fixed second pulse is chosen on 20 validation cells. Bayes
minimizes expected final variant entropy from the observed current, with 64
noise draws per hypothesis and action, and does not see the regime label.
Evaluation uses 200 held-out draws (50 per hypothesis), plus a doubled-noise
shift.

\begin{table}[H]
\centering
\caption{Crossed task on 200 simulated draws from four fixed hypotheses, not
biological replicates. Entropy is nats and is lower when the experiment is
informative. At 800~pA the learned values span four seeds.}
\label{tab:crossed}
\footnotesize
\setlength{\tabcolsep}{4pt}
\begin{tabular}{lrr}
\toprule
Controller & 400~pA $\downarrow$ & 800~pA $\downarrow$ \\
\midrule
Validation-selected fixed & 0.2844 & 0.5366 \\
Learned design (four seeds) & 0.1992 & 0.4887--0.4912 \\
Bayesian contingent design & 0.1992 & 0.4887 \\
Learned, current masked & 0.2844 & 0.5366 \\
\bottomrule
\end{tabular}
\end{table}

At nominal noise, all four learned seeds and Bayes choose pulse 0 on all 100
regime-0 draws and pulse 1 on all 100 regime-1 draws. Masking makes every
learned seed choose pulse 1, matching the fixed protocol. The paired
fixed-minus-learned entropy improvement is 0.0852 nats (95\% interval
0.0632--0.1086). Bayes ties the learned policies at this noise. At 800~pA the
learned seeds make 6, 4, 1, and 4 nonstandard regime-1 choices; Bayes makes
none. A prediction waveform was chosen after the first entropy result, so those
scores are exploratory. On a fresh 200-draw block the learned-minus-fixed gain
on the excluded waveform is $4.6991\times10^{-6}$ nats per sample (interval
$[2.49,7.20]\times10^{-6}$).

\begin{figure}[t]
\centering
\includegraphics[width=\linewidth,height=1.55in,keepaspectratio]{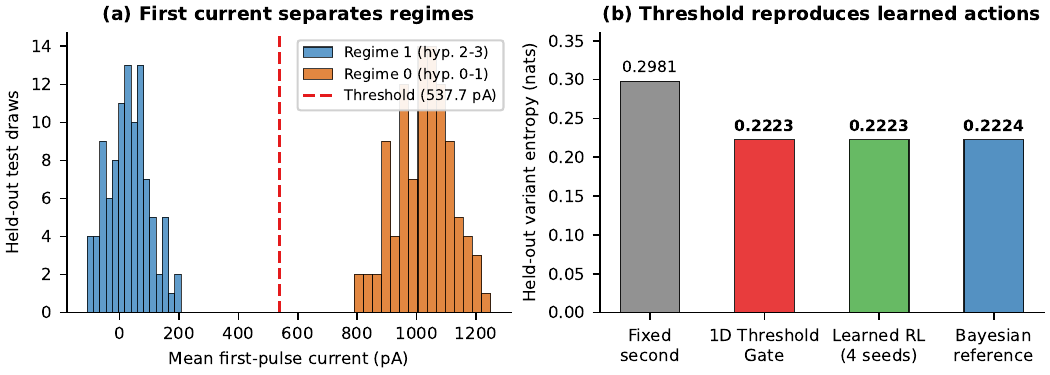}
\caption{(a) Mean first-pulse current on seeds 5000--5199 around the
validation-fit threshold ($537.7$~pA). (b) Held-out variant entropy. The
threshold reproduces all learned actions on this test block; its entropy is
nearly identical to the Bayesian reference, which agrees on 199 of 200 actions.}
\label{fig:threshold_gate}
\end{figure}

A fitted one-threshold rule reproduces the learned policies on the evaluated
block (Figure~\ref{fig:threshold_gate}). Finite agreement does not make the
threshold the Bayesian controller. The rule was fit only on seeds 4000--4199,
minimizing mean variant entropy over thresholds on the mean first-pulse current.
Current above 537.7~pA selects pulse 0 and lower current selects pulse 1. On
seeds 5000--5199 the frozen rule matches all four learned policies on every
draw. Both score 0.2223 nats, against 0.2981 for the fixed pulse. Bayes agrees
on 199 of 200 draws and scores 0.2224 nats. Their paired difference is
$4.39\times10^{-6}$ nats, and the interval includes zero.

\section{Implications and limitations}
\begin{table}[H]
\centering
\caption{Three questions, answered separately. Crossed-task entropy is the
training objective. The independent outcome is the fresh excluded waveform.
The simpler references are Bayes and the fitted threshold.}
\label{tab:questions}
\scriptsize
\setlength{\tabcolsep}{3pt}
\begin{tabular}{@{}p{1.35in}p{1.95in}p{1.95in}@{}}
\toprule
Question & Original task & Crossed task \\
\midrule
\raggedright Does feedback change actions? &
\raggedright No. All 1,200 rollouts are LL. Masking changes none. &
\raggedright Yes at nominal noise: the second pulse follows the first current. \tabularnewline
\raggedright Does it improve an independent outcome? &
\raggedright No. Learned and fixed scores coincide. &
\raggedright Fresh excluded-waveform gain: $4.6991\times10^{-6}$ nats/sample. \tabularnewline
\raggedright Does learning beat a simpler adaptive choice? &
\raggedright No. It copies the training-selected sequence. &
\raggedright No. It ties Bayes. The threshold reproduces learned actions; Bayes agrees on 199 of 200 and is not that rule. \tabularnewline
\bottomrule
\end{tabular}
\end{table}

Table~\ref{tab:questions} is the comparison. The original policies had limited
training on a fixed cell, so ignoring feedback there does not show that every
well-trained recurrent policy would do the same. Recording durations remain
8, 18, or 28~ms on that task. The crossed task changes the second pulse, but
only as one choice after a fixed diagnostic pulse, and its prediction gain is
$4.6991\times10^{-6}$ nats per sample. A fitted one-threshold rule reproduces
the learned policies on seeds 5000--5199; it is not the Bayesian controller.
Neither task shows general RL superiority.

Both tasks are synthetic, and the original scorer knows the apparatus. The
crossed task omits clamp artifacts. Intervals describe simulated draws from
four fixed hypotheses, not biological replication. The three failed pilots, the
tiny prediction gain, and the threshold split on mean first-pulse current are
part of the evidence, not a richer feedback policy. Evaluations of scientific
agents should include fixed controls, finite-horizon planning, excluded
targets, recording time, and observation ablations.

\end{document}

%% file: results_table.tex
Fixed (training-selected) & 0.0407 & 19.17 & 28.00 \\
Random & 0.0401 & 19.44 & 17.00 \\
Ideal greedy & 0.0370 & 20.88 & 9.08 \\
Ideal lookahead & 0.0370 & 20.88 & 9.08 \\
Oracle greedy & 0.0451 & 19.97 & 20.54 \\
Oracle lookahead & 0.0455 & 19.87 & 20.29 \\
Oracle predictive & 0.0458 & 20.19 & 18.75 \\
Recurrent policy & 0.0407 & 19.17 & 28.00 \\
Policy, current masked & 0.0407 & 19.17 & 28.00 \\
\bottomrule